\documentclass[conference]{IEEEtran}
\usepackage{times}

\usepackage[numbers]{natbib}
\usepackage{multicol}
\usepackage[bookmarks=true]{hyperref}
\usepackage{graphicx}
\usepackage{xcolor}

\usepackage{algorithm}
\usepackage[noend]{algorithmic}
\usepackage{wrapfig}
\usepackage{amssymb}
\usepackage[normalem]{ulem}
\usepackage{mathtools}

\usepackage{lipsum}

\begin{document}

\title{
Play it by Ear: Learning Skills amidst Occlusion through Audio-Visual Imitation Learning
}

\author{Maximilian Du$^*$, Olivia Y. Lee$^*$, Suraj Nair, Chelsea Finn \\
        Stanford University \\
        \texttt{\{maxjdu, oliviayl, surajn\}@stanford.edu} \\
  \scriptsize{* equal contribution}  }

\maketitle

\begin{abstract}
Humans are capable of completing a range of challenging manipulation tasks that require reasoning jointly over modalities such as vision, touch, and sound. Moreover, many such tasks are \emph{partially-observed};
for example, taking a notebook out of a backpack will lead to visual occlusion and require reasoning over the history of audio or tactile information. While robust tactile sensing can be costly to capture on robots, microphones near or on a robot's gripper are a cheap and easy way to acquire audio feedback of contact events, which can be a surprisingly valuable data source for perception in the absence of vision.
Motivated by the potential for sound to mitigate visual occlusion, we aim to learn a set of challenging partially-observed manipulation tasks from visual and audio inputs. Our proposed system
learns these tasks by combining offline imitation learning from a modest number of tele-operated demonstrations and online finetuning using human provided interventions. In a set of simulated tasks, we find that our system benefits from using audio, and that by using online interventions we are able to improve the success rate of offline imitation learning by $\sim$20\%. 
Finally, we find that our system can complete a set of challenging, partially-observed tasks on a Franka Emika Panda robot, like extracting keys from a bag, with a 70\% success rate, 50\% higher than a policy that does not use audio. 
\end{abstract}

\IEEEpeerreviewmaketitle

\section{Introduction}

\begin{figure}[t]
    \centering
    \includegraphics[width=0.99\linewidth]{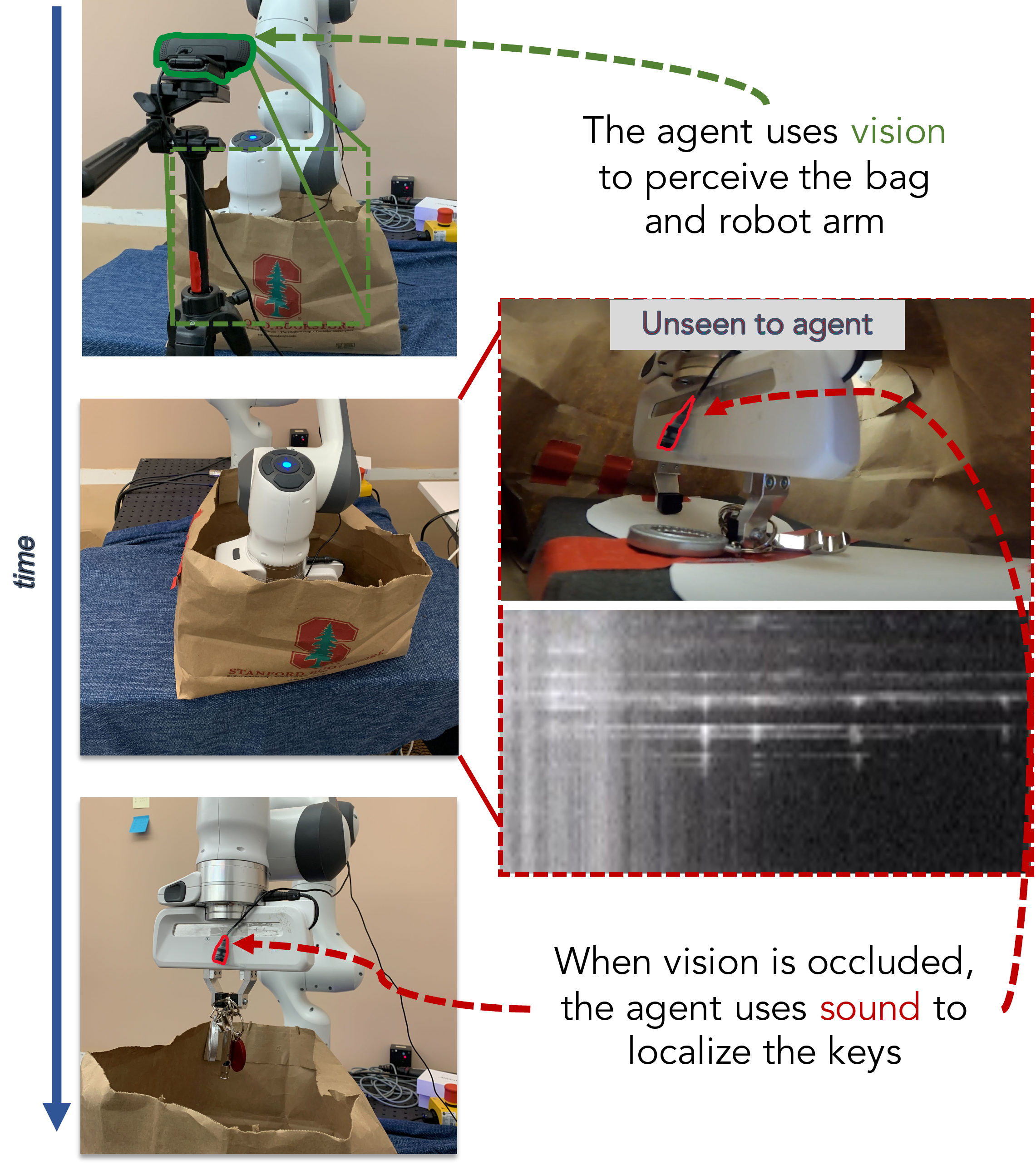}
    \vspace{-0.3cm}
    \caption{\small \textbf{Tackling Partially-Observable Tasks with Audio-Visual Imitation}. We tackle vision-based manipulation tasks with occlusions via interactive imitation learning where the agent leverages vision and audio modalities. For the task of extracting keys from a bag, the robot uses vision to perceive the bag, then uses sound from a microphone located on the gripper to detect the keys to grasp. The robot uses memory to encode the history of visual and audio observations to successfully extract the keys. 
    }
    \vspace{-0.6cm}
    \label{pullfig}
\end{figure}

End-to-end learning based approaches have shown impressive results when applied to vision-based robotic manipulation across a range of skills like grasping \cite{pinto2016supersizing, kalashnikov2018qtopt, levine2018learning} and manipulation of both rigid and deformable objects \cite{levinee2e, ebert2018visual, Hoque2021VisuoSpatialFF}. 
However, one pervasive yet often overlooked issue in vision-based manipulation is \textit{occlusion},
which can lead to partial observability and in some cases can make it impossible to solve the task altogether. Consider for example the task of extracting keys from a grocery bag.
While vision is necessary to successfully reach within the bag, the keys themselves are out of sight. Therefore, it is impossible to complete the task with vision only. One way to tackle this issue is by introducing additional sensor modalities like tactile feedback, which can reduce partial observability \cite{Lee2019MakingSO}, but accurate and robust tactile sensing for robot learning remains an open problem.
We instead consider \emph{audio feedback} as an alternative, since microphones are both cheap and robust, and as we will find, a microphone attached to the robot's gripper can provide sufficient information about contact when camera images are occluded.
Therefore, in this work 
we focus on the problem of learning robotic manipulation tasks amidst occlusion via human demonstrations, using vision and audio modalities.

While bringing in additional sensor modalities beyond vision can help mitigate occlusion, a number of challenges remain: most notably, such problems still involve partial observability. Even if an agent captures multi-modal observations (i.e. both vision and audio), each sensor may provide relevant information intermittently over time, and the agent must reason about the relationships between different sensor modalities at different times.
For example, in the task of extracting the keys from the cloth bag, a sound made when the gripper hits the keys many timesteps in the past may be critical in identifying the keys' position and successfully grasping them. Thus, our agent leverages memory to learn from a history of audio and visual observations.

Learning from multiple modalities over an extended history presents yet another challenge: it produces a large and temporally correlated observation space. This is a major challenge for agents trained with only offline imitation learning,
as it is easy for the learned agent to veer off the expert state distribution and fail on out-of-distribution states. This state distribution shift challenge is widely studied in imitation learning \cite{Ross2011ARO}, and is exacerbated
when the agent considers a history of observations.
For example, an agent with memory can be at a state within the expert state distribution, but if the trajectory it took to get there is not, then the input to the policy may still be out-of-distribution.
To mitigate this, we adopt the well-studied strategy of using \emph{online} expert feedback \cite{Ross2011ARO, Kelly2019HGDAggerII} to finetune and correct the agent's behavior. 

Towards tackling these problems, we propose and develop a system
with three key properties. First, to handle visual occlusion, we learn from a multi-modal sensor input containing both third-person RGB images and audio data from a microphone attached to the robot's gripper. Second, to address partial observability, our system leverages LSTM \cite{lstm} memory
to capture the history of audio and visual observations and their interactions. These are all encoded in an end-to-end network which outputs the robot's end-effector actions. Finally, to handle the state-distribution shift from learning with offline demonstrations in this large observation space, we use online corrections from an expert (i.e. interventions) to finetune the learned policy in a sample efficient way.

Concretely, our main contribution is an end-to-end imitation learning system for completing partially-observed vision-based manipulation tasks amidst occlusion by leveraging audiovisual sensor modalities, memory, and interactive learning from expert corrections. We conduct extensive experiments in simulation evaluating the importance of design choices around fusing modalities, leveraging memory, and online finetuning for task completion. Finally, we find that our system
is capable of completing multi-modal, partially-observed tasks (e.g. extracting keys from a bag) entirely from vision and sound on a real Franka Emika Panda robot (See Figure \ref{pullfig}) with a 70\% success rate, more than double the success rate of an agent using only vision.

\section{Related Work}

Towards learning to completing partially-observed manipulation tasks, our approach draws on ideas from multi-modal learning, learning with partial observability, and efficient finetuning of trained policies.

\vspace{0.1cm}\noindent\textbf{Multi-Modal Robotic Learning.}
Learning from multiple sensor modalities is a well-studied problem in robotics. For example, a number of works have studied how to effectively fuse information from visual and tactile input to enable better reinforcement learning \cite{hoof_autoencode_2016, calandra_grasp_2018, Lee2019MakingSO, Kumar2019ContextualRL, Ichiwara2021ContactRichMO} and imitation learning \cite{Liu2020UnderstandingMP, Balakuntala2021LearningMC},
as well as to identify and reconstruct object properties \cite{Li2019ConnectingTA, Taunyazov2020EventDrivenVS, Smith20203DSR, Zhao2021ANA}. Like these works, we aim to leverage multiple modalities to complete robotic manipulation tasks, but unlike these works, use vision and \emph{audio} modalities.

A number of works have also studied multi-modality from vision and sound in the context of visual representation learning \cite{Arandjelovi2018ObjectsTS, Morgado2020LearningRF, Gao2021ObjectFolderAD}, sound synthesis \cite{Owens2016VisuallyIS}, and robotics \cite{Okuno2002, wu2009, clarke2019robot, Dean2020SeeHE, Chen2020AudioVisualWF, Gandhi2020SwooshRT, Gan2020NoisyAS, chen2021structure}.
In robotics, these approaches have ranged from using vision and sound for exploration in robotic navigation \cite{Dean2020SeeHE,Gan2020NoisyAS, Chen2020AudioVisualWF, chen2021structure} to using sound to identify object properties and robot actions
\cite{Gandhi2020SwooshRT}. 
Our work instead focuses on using it as an input modality for end-to-end learning of manipulation tasks. 
Like our work, \citet{clarke2019robot} uses sound as input for manipulation, specifically to learn robotic scooping behavior, while we instead focuses on fusing vision and sound within closed-loop control, necessary for completing tasks that contain occlusion.

\vspace{0.1cm}\noindent\textbf{Learning with Partial Observability.} Partial observability in reinforcement and imitation learning is well-studied in the context of robotics problems like egocentric navigation. Many works use an LSTM \cite{lstm} augmented agent for tasks like room or target-driven navigation \cite{Mirowski2017LearningTN, Wu2018BuildingGA, Mousavian2019VisualRF}. Other works have used more sophisticated memory architectures and Transformers \cite{Vaswani2017AttentionIA} to handle partial observability in navigation tasks \cite{Oh2016ControlOM, Parisotto2018NeuralMS, fang2019smt, Parisotto2020StabilizingTF}. While these works primarily focus on navigation and partial observability from ego-centric motion, our work instead focuses on tackling occlusion in vision-based manipulation and employs memory in combination with multi-modal learning from vision and sound. 

RNN augmented agents \cite{Mandlekar2021WhatMI} and structured scene representations \cite{Xu2020Learning3D} have seen some promising results in the context of manipulation and occlusions. 
However, in this work we consider the case where the target object is always fully occluded, and only incorporating additional sensor modalities can resolve the partial observability. 

\vspace{0.1cm}\noindent\textbf{Efficient Finetuning of Learned Policies.} Critical to our method's performance is the ability to improve the policy trained from offline demonstrations through online interaction. A natural approach to this is to use reinforcement learning to finetune the policy \cite{Vecerk2017LeveragingDF, pmlr-v80-kang18a, Julian2020EfficientAF, Nair2020AcceleratingOR, Lu2021AWOptLR} or train a residual policy \cite{Johannink2019ResidualRL, Alakuijala2021ResidualRL}. One drawback of such methods is that they require a reward function which can be difficult to provide on a real robot, and as we see in Section~\ref{sec-exp}, can still be highly sample-inefficient. Instead, we adopt the paradigm of interactive imitation learning \cite{Ross2011ARO}, where an online expert can provide a modest number of targeted interventions to the previously trained agent to correct their behavior, either initiated by the human \cite{Kelly2019HGDAggerII, spencer_interventions, Mandlekar2020HumanintheLoopIL, jang2021bcz} or by the robot \cite{Zhang2016QueryEfficientIL, Hoque2021LazyDAggerRC, Hoque2021ThriftyDAggerBN}. We incorporate components of human-gated DAgger (HG-DAgger) \cite{Kelly2019HGDAggerII}
where the human intervenes when necessary, and the policy is finetuned on the interventions and original offline demonstrations.

\begin{figure*}[t]
    \centering
    \includegraphics[width=0.99\linewidth]{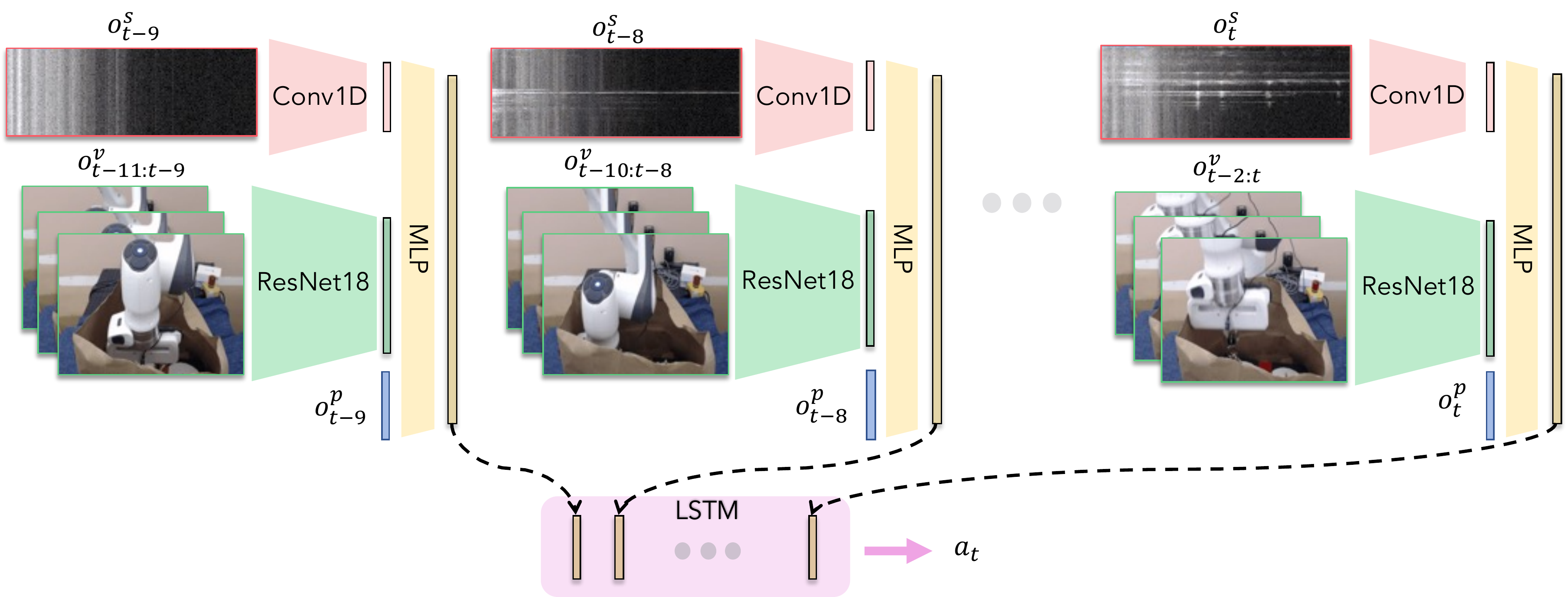}
    \caption{\small \textbf{Our Proposed Architecture for Encoding Vision and Audio}. In order to capture vision, sound, and their inter-dependencies over time, we propose the above architecture. At each timestep we encode the immediate history of visual and audio feedback (last 3 frames of images, last 2 seconds of audio) using a ResNet18 visual encoder and a spectrogram + 1D convolution respectively. These visual and audio encodings are concatenated with proprioceptive data, fused through an MLP, and fed into an LSTM over the last 10 timesteps to capture the long-range temporal dependencies between the multi-modal inputs, finally outputting the current end-effector action.}
    \vspace{-0.6cm}
    \label{methodfig}
\end{figure*}

\vspace{0.1cm}\noindent\textbf{Object Extraction and Occlusion in Robotic Manipulation.}
Finally, there is a rich literature of works in robotics that specifically aim to tackle tasks with occlusion. For example, a number of techniques tackle tracking and pose-estimation of either the robot gripper \cite{cifuentes_2017} or a target object \cite{cheng_2019, wen_2020, luo_2021} in the presence of occlusions. Other works build articulated 3D models of objects \cite{huang_2012}, allowing them to be manipulated even if areas of the object are occluded. Unlike these works, we focus on tasks in which the target is never observed visually, and other modalities like sound are needed to localize it.

Other works have tackled the problem of manipulating objects in clutter, also known as mechanical search \cite{mechsearch, xray, Bejjani2021OcclusionAwareSF}. While these works do deal with fully occluded objects, they assume the object is visible after the objects on top of it have been removed, an assumption which does not apply in the case of extracting an object from within a bag. 

Finally, other approaches to manipulation in full occlusion involve using radio frequency (RF) perception \cite{Boroushaki_2021} or jointly learning to manipulate and move the agents camera \cite{pmlr-v87-cheng18a}. Unlike RF perception, our method uses relatively cheap and accessible microphones, and does not require the target object to have an RFID tag. While learning to move the camera can help in some cases of occlusion, it can make learning more challenging and even in the optimal position, a camera may provide little useful information inside of a poorly-lit bag or backpack.

\section{Memory Augmented Audio-Visual Imitation} 
\label{sec-method}

Towards tackling tasks with heavy visual occlusion, we present a system for imitation learning from visual and audio modalities that brings together three key components. 
\textbf{First}, we leverage an end-to-end learned approach that takes as input and fuses both vision and sound modalities. \textbf{Second}, we use a memory augmented neural network that captures the multi-modal inputs over an extended history.
\textbf{Finally}, we leverage HG-DAgger \cite{Kelly2019HGDAggerII}, an interactive imitation learning approach for finetuning the policy trained on offline demonstrations efficiently given online interaction. As we find in Sections~\ref{sec:robot-res} and ~\ref{sec:e2}, each of these components is critical to the performance of the complete system.

\subsection{Preliminaries}

Before describing each component of our system in detail, we define our problem setup and notation. 
We assume that the agent operates in a controlled Markov process $\mathcal{M} = \{\mathcal{S}, \mathcal{A}, p, \mu, T\}$ where $\mathcal{S}$ is the state space, $\mathcal{A}$ is the agent's action space, $ p(s_{t+1} \vert s, a)$ is the transition distribution, $\mu(s_0)$ is the initial state distribution, and $T$ is the episodes' horizon. The agent does not receive true states $s \in \mathcal{S}$, but rather receives observations $\mathcal{O}$ that are partially observed (i.e. the observation space is non-Markovian).
We aim to learn a task given a dataset of $N$ expert demonstrations $\mathcal{D} = [(o_0, a_0), (o_1, a_1), ..., (o_T, a_T)]_{1:N}$ consisting of the agents observations $o \in \mathcal{O}$ and actions $a \in \mathcal{A}$.

Because our goal is to handle tasks with heavy occlusion, we consider a multi-modal observation space, where each observation consists of components $o = [o^v, o^s, o^p]$, where $o^v \in \mathcal{O}^v$ is the agent's visual observation, $o^s \in \mathcal{O}^s$
is the agent's audio observation, and $o^p \in \mathcal{O}^p$ is the agent's proprioceptive observation (e.g. end-effector position and gripper position). 
Our goal then is to learn a parametric policy with memory $\pi_\theta: \mathcal{O}^v_{(t-H):t} \times \mathcal{O}^s_{(t-H):t} \times \mathcal{O}^p_{(t-H):t} \rightarrow \mathcal{A}$.
The policy parameters $\theta$ are
trained to minimize
a conventional mean-squared error behavior cloning loss, that is:
\begin{equation}
    \mathcal{L}_{bc}(o_{(t-H):t}, a_t) = ||a_t - \pi_{\theta}(o_{(t-H):t})||_2^2
    \label{eq:bcloss}
\end{equation}

\subsection{Learning from Vision, Sound, and Proprioception}

To tackle tasks with full occlusion, a key component of our system is the ability to fuse inputs coming from visual, audio, and proprioceptive modalities (See Figure~\ref{methodfig}).

\vspace{0.1cm}\noindent\textbf{Vision.} We record images with an uncalibrated third-person camera mounted in front of the robot, viewing the robot arm and scene (though the target object itself is fully obscured from view of the camera).
We capture these visual observations $o^v$ 
as 84x84 RGB images.
We pre-process the images by upsampling them  to 224x224, normalizing them, and then feeding them through a convolutional neural network encoder, specifically a ResNet18 \cite{He2016DeepRL} which produces a 512 dimensional image embedding $e^v$. During training, the images from the dataset are augmented using random cropping, color jitter, and random affine transformations, which have been shown to be effective for more robust imitation and reinforcement learning \cite{Srinivas2020CURLCU, Mandlekar2021WhatMI}. 

\vspace{0.1cm}\noindent\textbf{Sound.} For audio observations $o^s$, we capture the last two seconds of the audio waveform from a microphone attached to the gripper. Because the raw waveform is high dimensional and noisy, we preprocess it through a spectrogram as done in prior work \cite{clarke2019robot}.
Specifically, the output of the spectrogram is of shape $[L, S]$ where $L$ are discrete timesteps (in our case $L = 57$) 
and $S$ are the different frequency bins of the spectrogram (in our case $S = 160$). 
To capture the temporal equivariance in the output of the spectrogram, we then pass the $[57, 160]$ matrix through a 1D convolutional neural network, producing an output of size $[8, 33]$, which is flattened to produce the final audio embedding $e^s$ of size $264$.

\vspace{0.1cm}\noindent\textbf{Proprioception.} We record the robot's internal proprioceptive data, that is, measurements of its own end-effector position and gripper status, yielding an observation $o^p$ of shape $(1, 7)$.

Finally, the embeddings of the visual input, audio input, and the proprioceptive features $[e^v, e^s, o^p]$ are concatenated to form the observation embedding that is used to predict action. However, as we will find in our experiments, simply using single timestep information leads to poor performance, due to the inherent partial observability in the tasks we consider. We address this issue by encoding history, which we cover next.

\begin{figure}[t]
    \centering
    \includegraphics[width=0.9\linewidth]{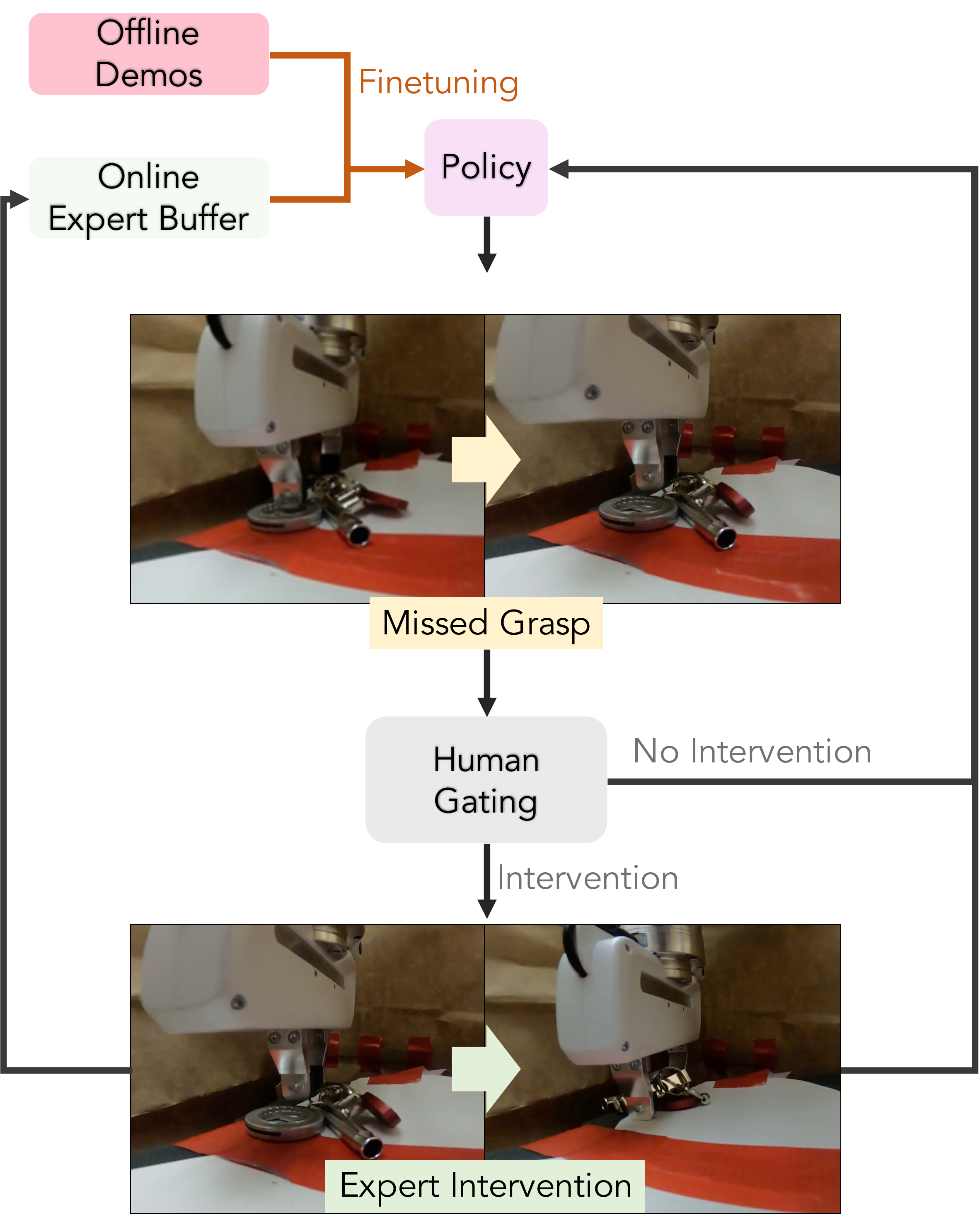}
    \vspace{-0.3cm}
    \caption{\small \textbf{Online Finetuning via Interactive Imitation}. The multi-modal and high-dimensional nature of our task means purely offline BC can suffer from state-distribution shift. To improve the robot's performance, we leverage HG-DAgger \cite{Kelly2019HGDAggerII}, where a human expert is in the loop of learning, and intervenes when necessary. The policy is finetuned on balanced batches of the original offline demonstrations and the online actions provided by the expert. }
    \vspace{-0.6cm}
    \label{method_interactive}
\end{figure}

\subsection{Encoding History}

To capture the intermittent information between the different sensor modalities, the second component of our system focuses on how to encode and fuse the history of multi-modal observations. To do so we perform
imitation learning with a recurrent memory \cite{Rahmatizadeh2016LearningMT}. 
Since the tasks we are interested in may require reasoning over short timescales (for example the motion of the robot) as well as longer timescales (the audio signal from several past timesteps), we opt to combine both early and late fusion across time (See Figure~\ref{methodfig}).

\vspace{0.1cm}\noindent\textbf{Early Fusion.}
Effectively manipulating objects in the presence of occlusion requires perception of the motion of the arm and the objects in the scene;
thus we provide the agent with a temporal signal of the motion of the arm over short time sequences by using frame-stacking.
Specifically, we stack the past 3 timesteps of observations $[o^v_{t-2}, o^v_{t-1}, o^v_{t}]$ channel-wise, and feed them into a modified ResNet18 with 9 input channels to produce the visual embedding $e^v_t$.

\vspace{0.1cm}\noindent\textbf{Late Fusion.} Since the robot will need
to combine audio and visual inputs at different times to successfully complete the task, we also fuse all encoded modalities over time. Specifically, we concatenate the encoded visual, audio, and proprioceptive embeddings $[e^v, e^s, o^p]_t$ at every step $t$ then feed them into an MLP that creates a joint embedding $z_t$. We pass in a sequence of these $z_t$ embeddings across the past $H$ timesteps into a one-layer LSTM, taking the final $D$ dimensional encoding. 
In our implementation $H$ is 10 and $D$ is $50$. 
This encoding is fed into an MLP with two hidden layers, which finally outputs the continuous delta end-effector action and discrete gripper action. Please see the appendix for further implementation details.

\begin{figure*}[t]
    \centering
    \includegraphics[width=0.9\linewidth]{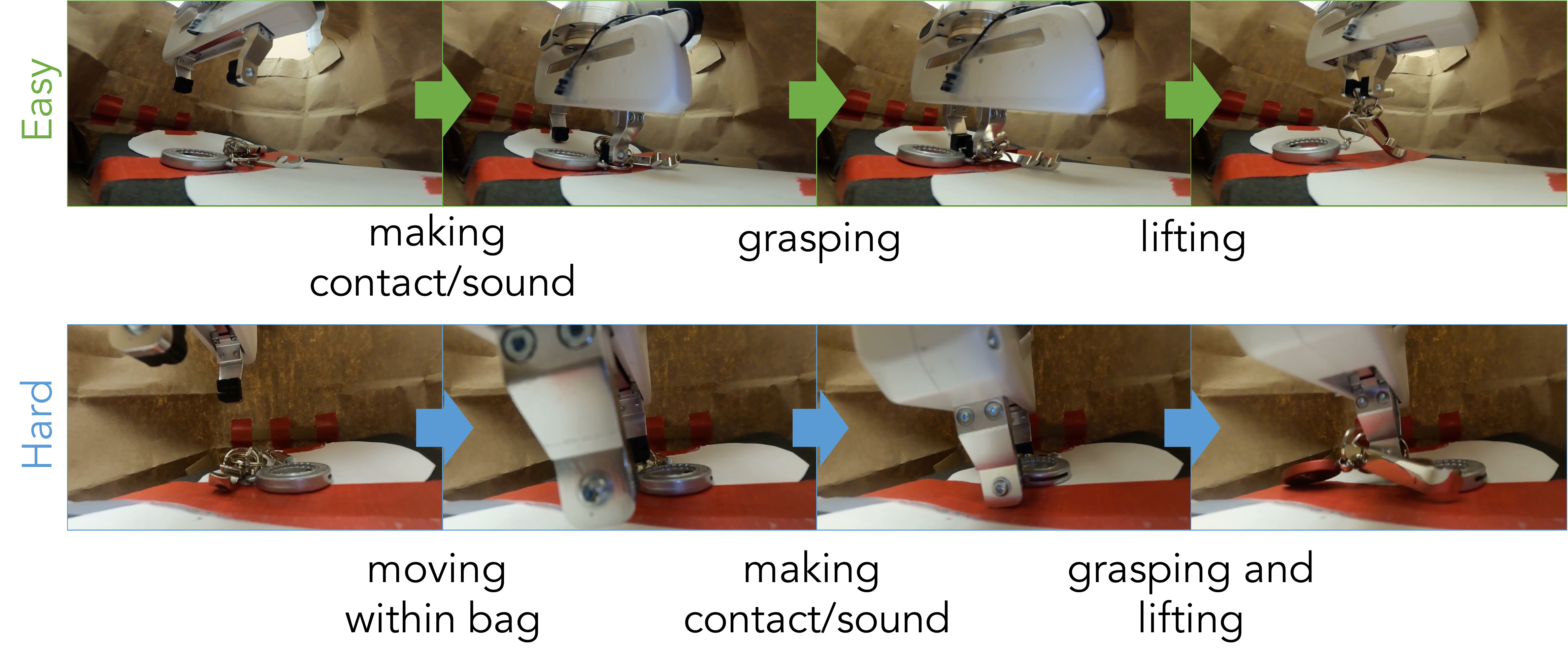}
    \vspace{-0.3cm}
    \caption{\small \textbf{Qualitative Examples of Task Success}. In the Easy version of the task the agent moves directly down until it hears contact with the keys, at which point it can grasp and extract them. In the Hard task, the agent must move around the base of the bag until it hears the keys move, then can grasp and lift the keys. \emph{Note: This camera view is not visible to the agent.} }
    \vspace{-0.3cm}
    \label{robot_qual}
\end{figure*}

\begin{algorithm}[t]
\caption{Our Full System}
\begin{algorithmic}[1]
\STATE \textbf{Input}: Expert demos $\mathcal{D} = [(o_1, a_1), ..., (o_T, a_T)]_{1:N}$
\STATE Initialize $\mathcal{D}_{o} \leftarrow \emptyset$, Initialize $\theta$ randomly
\STATE \textcolor{blue}{/* Train on offline demos until convergence */}
\STATE Repeatedly update $\pi_\theta$ according to Eq.~\ref{eq:bcloss}
\FOR{$ep=1, 2, ..., E$}
    \WHILE{episode not done}
    \IF{human intervenes}
    \STATE \textcolor{blue}{/* Step expert action and store it */}
    \STATE $a^*_t = \pi_{human}(o_{(t-H):t})$
    \STATE $\mathcal{D}_{o} \leftarrow \mathcal{D}_{o} ~\bigcup~ (o_{(t-H):t}, a^*_t)$
    \STATE $s_{t+1} \sim p(\cdot \mid s_t, a^*_t)$
    \ELSE 
    \STATE \textcolor{blue}{/* Step policy action */}
    \STATE $a_t = \pi_{\theta}(o_{(t-H):t})$
    \STATE $s_{t+1} \sim p(\cdot \mid s_t, a_t)$
    \ENDIF
    \ENDWHILE
    \FOR{num updates $U$}
        \STATE \textcolor{blue}{/* Finetune on balanced batches  */}
        \STATE $o_{(t-h):t}, a_{t} \sim \mathcal{D}$
        \STATE $o^*_{(t-h):t}, a^*_{t} \sim \mathcal{D}_{o}$
        \STATE Update $\pi_\theta$ according to Eq.~\ref{eq:bcloss_online}.
    \ENDFOR
\ENDFOR
\end{algorithmic}
\end{algorithm}

\subsection{Online Finetuning via Interventions}
\label{sec-method-interactive}

Finally, even with learning from multiple modalities and memory, offline imitation learning still can face challenges with state-distribution shift during rollouts. To address this, the third component of our approach is to fine-tune
the recurrent audio-visual policy trained on offline demonstrations through online interactive imitation learning. Specifically, we incorporate the approach of \citet{Kelly2019HGDAggerII}, where a human observes the robot interacting in the environment, and chooses when to intervene and take actions for the robot (See Figure~\ref{method_interactive}).

Specifically, the policy $\pi_\theta$ is first trained offline on a set of expert demonstrations $\mathcal{D}$ until convergence. Then, during online interaction, the human may take over and choose to intervene, and provide an expert action $a^*_t = \pi_{human}(o_{(t-H):t})$. These observation and action tuples from the human $(o_{(t-H):t}, a^*_t)$ are then stored in a separate online replay buffer $\mathcal{D}_{o}$. During finetuning, the policy $\pi_\theta$ is then updated to minimize the behavior cloning loss on balanced batches from $\mathcal{D}_{o}$ and $\mathcal{D}$:
\begin{equation}
\begin{split}
    \mathcal{L}_{ft} = \mathbb{E}_{o, a \sim  \mathcal{D}}[\mathcal{L}_{bc}(o_{(t-H):t},a)] +\\ \mathbb{E}_{ o, a^* \sim \mathcal{D}_{o}}[\mathcal{L}_{bc} (o_{(t-H):t}, a^*)]
    \label{eq:bcloss_online}
\end{split}
\end{equation}
Our full system is summarized in Algorithm 1.

\section{Experiments}
\label{sec-exp}
In our experiments, we aim to study the efficacy of this system on fully-occluded manipulation tasks in both simulation and on a real robot.
We begin in Sections~\ref{sec:robot-env} and~\ref{sec:robot-res} where we describe our physical robot environment and results, which include a Franka Emika Panda robot extracting keys from a bag. 
We observe that our approach is able to complete this task with a 70\% success rate, while a pure vision-based approach only gets a 20\% success rate. Furthermore, we see our method exhibits some generalization to unseen objects. In Section~\ref{sec:robotsound}, we describe our second real robot experiment which illustrates our methods ability to differentiate between the sounds of different objects.
Then in Section~\ref{sec:sim}, we describe our simulation environments and tasks, where we conduct extensive experiments studying the different components of our system. Finally, in Section~\ref{sec:e2}, we ablate different design choices of our method in simulation, and again observe that removing online learning, either modality, or memory hurts performance. Please see \url{https://sites.google.com/view/playitbyear/} for videos of execution in simulation and on the robot.

\subsection{Real Robot Key Extraction Environment}
\label{sec:robot-env}

Motivated by our original goal of extracting keys from a bag, we aim to use our system to have our Franka Emika Panda robot extract a ring of keys from a shopping bag (See Figure~\ref{pullfig}). The robot receives 84x84 RGB camera observations from a third-person camera, as well as audio feedback from a microphone attached to the gripper. 

\begin{figure*}
    \centering
    \includegraphics[width=0.99\linewidth]{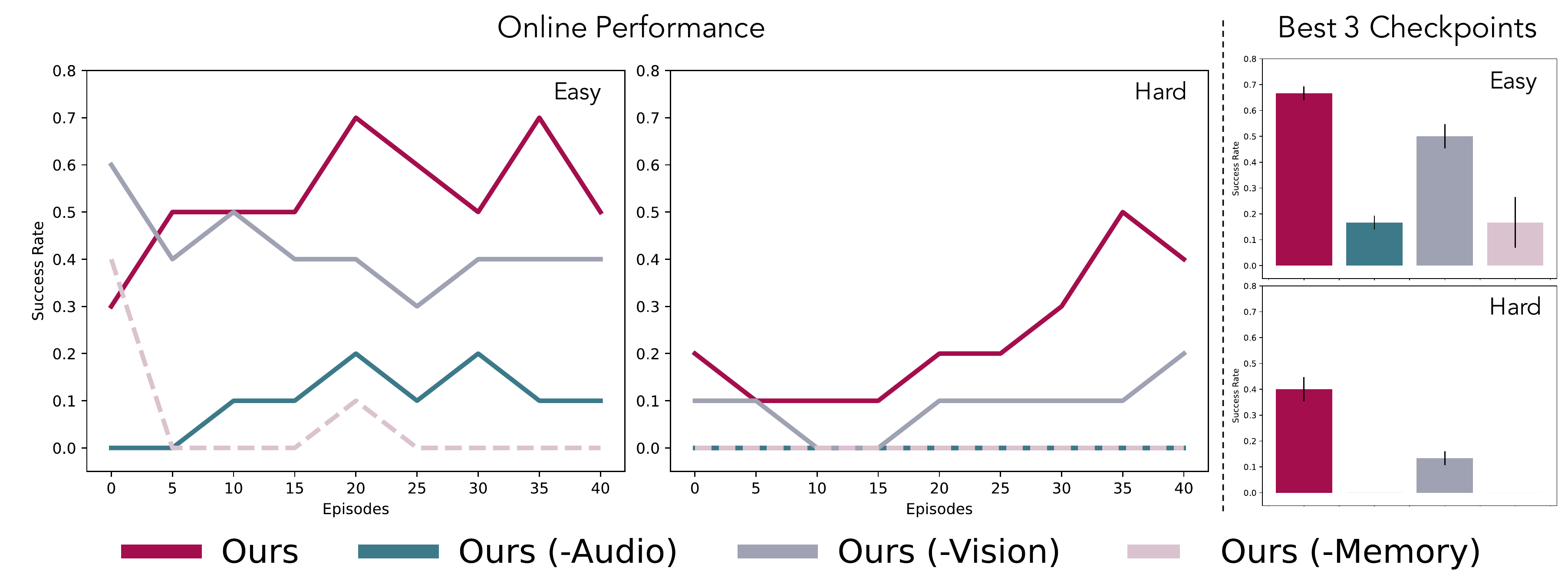}
    \vspace{-0.3cm}
    \caption{\small \textbf{Franka Key Extraction Success Rate}. We report the success rate of our policy on both the Easy and Hard versions of the key extraction task over the course of online finetuning (\textbf{left/middle}). We see that our system is able to get up to 70\% success rate on the easy version of the task with just 20 episodes of online interaction, and up to a 50\% success rate on the hard version of the task with just 40 episodes of online interaction. We also report the mean and standard error success rate of the best 3 checkpoints for each method (\textbf{right}). Overall we see that our full system performs significantly better than removing either audio, vision, or memory, and that for our full method online interventions are essential for strong final performance. }
    \vspace{-0.6cm}
    \label{exp3}
\end{figure*}

We consider two versions of the task, an \textbf{Easy} version where the keys are generally initialized in the center of the shopping bag below the gripper, and a \textbf{Hard} version where the keys are initialized over a wider radius of $\sim$3cm and in more challenging configurations. The primary difference is that in the easy version of the task, the robot can contact the keys and make a sound by moving directly down into the bag, while in the hard version, the robot must move around within the bag to make contact with the keys and generate a sound (See Figure~\ref{robot_qual}). Success is measured when the keys are lifted to a height such that they are visible to the camera. 

For our real robot task, we first collect 50 demonstrations using human tele-operation with an Oculus Quest 2 VR controller. The system described in Section~\ref{sec-method} is trained for 10,000 steps on the offline demos. Then, we rollout the policy online, with humans intervening as necessary as described in Section~\ref{sec-method-interactive}, finetuning the policy after each episode.
For all comparisons each method is finetuned online separately with interventions for a fair comparison.

\subsection{Real Robot Key Extraction Results}
\label{sec:robot-res}

We fine-tune our system and baselines over the course of 40 online episodes with human interventions where the first 12 episodes have the Easy initializations and the remaining 28 have Hard initializations.
Every 5 of these online intervention episodes, we evaluate the success rate of each method (without any interventions) on 10 trials of the Easy and 10 trials of the Hard tasks. We report the success rate of our system on each version of the task over online episodes in Figure~\ref{exp3}. Over the course of these 40 episodes we observe our method is able to accomplish a \textbf{70\% success rate on the Easy task and a 50\% success rate on the Hard task.}  Figure~\ref{robot_qual} shows sample rollouts of our system completing the Easy and Hard tasks.

First, we observe from Figure~\ref{exp3} that the Easy task starts with a 30\% success rate without any online finetuning (trained solely on offline demonstrations), and with just 20 episodes of online interaction ($\sim$1 hour of human effort), it is able achieve a success rate of up to 70\%. For the Hard task, learning takes longer, and the success rate hovers at $\sim$20\% for the first 25 episodes of online interaction. 
However, by the 30-episode mark
(once there are a sufficient number of online demonstrations on the harder task), 
the success rate begins increasing, and by 40 episodes of online intervention, the robot is able to complete the hard task with a $\sim$50\% success rate. These results suggest that online finetuning is critical to our system's performance, and just a 1-2 hours of human time can boost success significantly.

Next, we ablate the different design choices of our method. Specifically, we compare to \textbf{Ours (-Audio)} which removes any audio input, \textbf{Ours (-Vision)} which removes all visual input, and \textbf{Ours (-Memory)} which removes all memory from our method. All ablations are evaluated under the same protocol as our full system, and results can be found in Figure~\ref{exp3}.

For \textbf{Ours (-Audio)}, we observe that over all 40 online intervention episodes, the success rate hovers at $\sim$10\% for the \textbf{Easy} task (and never exceeds $\sim$20\% success rate), and the robot consistently failed on all the Hard tasks it was evaluated on. 
Qualitatively, we observe that for the Hard task, even when the agent did make contact with the keys and make sound, the robot would then move away from the keys. Furthermore, in both the Easy and Hard tasks, the agent often would attempt to grasp and lift even without making contact with the keys. The learned search policy of this agent was fixed, depending entirely on proprioceptive information, unlike our full method which moves until it hears sound, and only then grasps and extracts the keys. Based on these observations, we conclude that without audio, the robot is not able to identify when it has found the keys, or how to effectively move to localize them, and thus performs poorly at the task even with 40 episodes of online interaction.

For \textbf{Ours (-Vision)}, we observe that over 40 online intervention episodes, the success rate for the Easy task hovers around $\sim$50\% success rate, with the highest performance of 60\% coming without any interventions, we suspect due to noise in the evaluation. 
For the Hard task, the success rate remains at $\sim$10\% over nearly all 40 online intervention episodes. 
Qualitatively, we observe that on the Easy mode the agent can use sound and proprioception to move down until hearing the keys and grasping them. However on the Hard task, the robot needs to move around to make contact with the keys; when doing so, a common failure mode is when the robot engaged a false/premature lift resulting from a sound from contact with the grocery bag. We hypothesize that, because this ablation does not use vision, it cannot perceive the bag, it thus has a harder time differentiating between when noise comes from the gripper touching the bag vs. touching the keys. Our full system on the other hand does not exhibit this failure mode, as the bag deformation when the robot arm contacts the grocery bag is visible to the camera, enabling the robot to differentiate between the sound generated by touching the grocery bag and the sound generated by contacting the keys. This ablation suggests that, even with occlusions, incorporating whatever vision is available in addition to sound is important to the systems overall success rate.

Finally, we compare to \textbf{Ours (-Memory)}, which removes all memory from the system. Interestingly, on the Easy task, the policy trained without any interventions performs reasonably getting a 40\% success rate, quickly moving down and grasping and lifting. However once trained with interventions to show the agent to correct a missed grasp, the agent without memory struggles, since without memory it cannot perceive if it has successfully grasped or needs to regrasp. As a result it moves neither deeply nor quickly enough to produce sound and hovers around a $\sim10\%$ success rate. On the Hard task the agent always hovers around a $\sim10\%$ success rate for similar reasons, and even when it is able to make contact with and grasp the keys, it drops them prematurely, almost immediately after grasping them. We hypothesize that without the memory, it is unable to identify what stage of the task it is in, since it no longer encodes that it previously heard a sound and grasped the keys, and thus cannot lift the keys to complete the task.

Ultimately, we conclude that all of the ablations hurt the overall performance of our system on the real robot. A detailed quantitative analysis of the failure modes of our method and each ablation can be found in the appendix, along with further implementation and experiment details.   

\begin{table}[!h]
  \centering %
  \begin{tabular}{|c|c|c|c|c|c|} 
  
    \hline
    \multicolumn{1}{|c|}{Seen Objects} & \multicolumn{5}{|c|}{Unseen Objects} \\ 
    \hline  
    Keys & Aspirin Bottle & Chip Bag & Pens & Candies & Cloth \\
    \hline
    7/10 & 5/10 & 10/10 & 5/10 & 7/10 & 0/10 \\
    \hline

  \end{tabular}
  \caption{\textbf{Success Rate on Unseen Objects}. We test our policy trained on the keys \emph{zero-shot} on unseen objects. It is able to achieve some success on unseen objects \emph{that make noise}, like a bag of chips, aspirin bottle, or a pile of pens. Objects that don't make sound like a cloth don't succeed at all despite being relatively easy to grasp. }
  \label{tab1}
  \vspace{-0.6cm}
\end{table}

Next, we conduct an experiment on the real robot testing the trained policy's ability to generalize to new objects. We take the final policy of our full system trained with 40 online episodes, then evaluate it with different held out objects never seen (or heard) by the agent (See Table~\ref{tab1}). Interestingly, we observe that the robot is able to do a reasonable job grasping unseen objects like an aspirin bottle, bag of chips, or pile of pens, \textbf{as long as the object produces a loud enough sound upon contact}. Objects like a sponge or cloth which are relatively easier to grasp fail completely because they do not produce sound and the robot never closes its gripper.

\begin{figure}[t]
    \centering
    \includegraphics[width=0.99\linewidth]{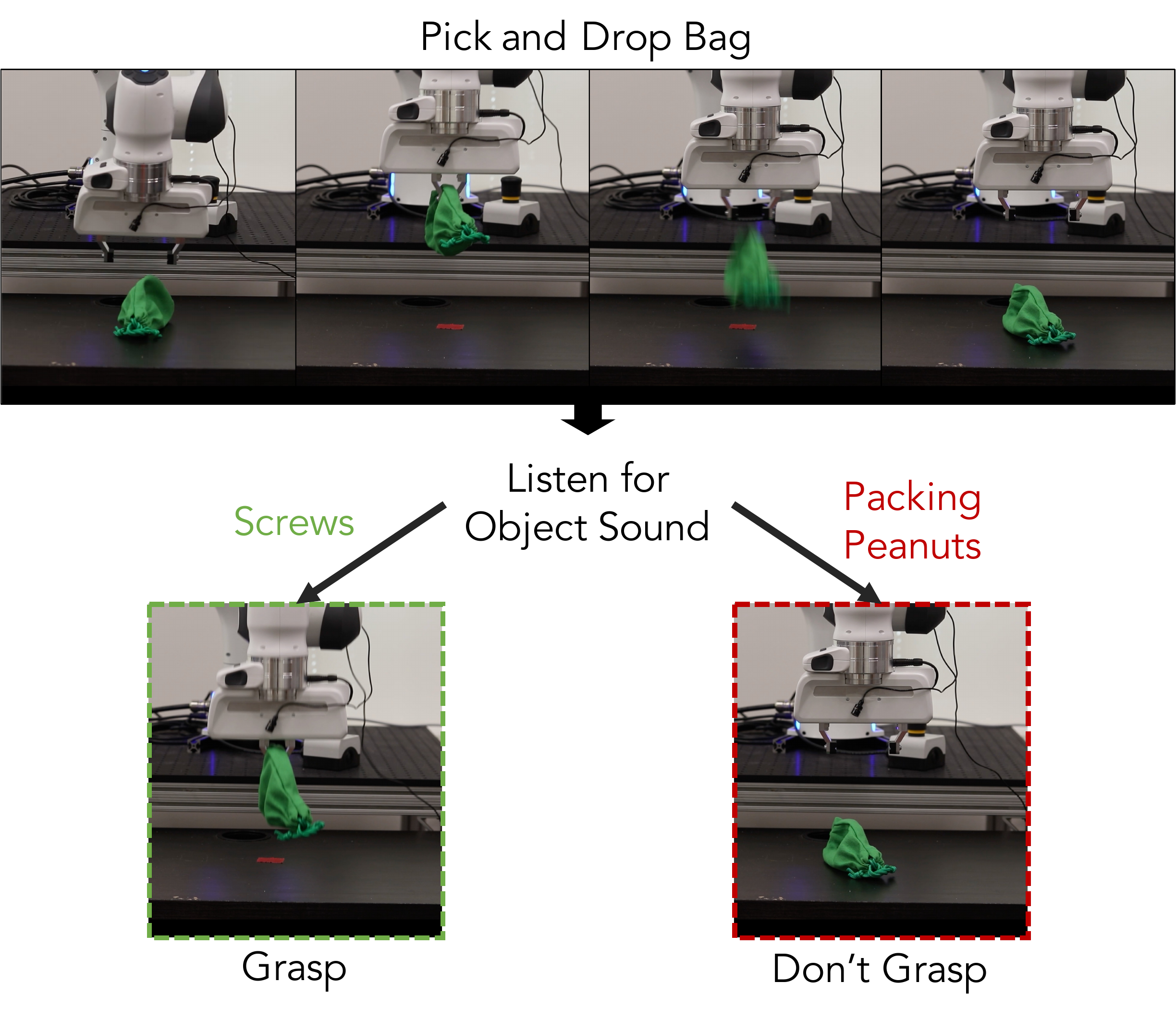}
    \vspace{-0.6cm}
    \caption{\small \textbf{Object Differentiation}. In our second robot experiment, we test our methods ability to differentiate between the sounds of different objects. In the task, the agent must pick and drop a mystery bag, and based on the sound of the object falling decide whether to grasp the bag or not.}
    \vspace{-0.6cm}
    \label{robotdomain2}
\end{figure}

\begin{figure*}[t]
    \centering
    \includegraphics[width=0.9\linewidth]{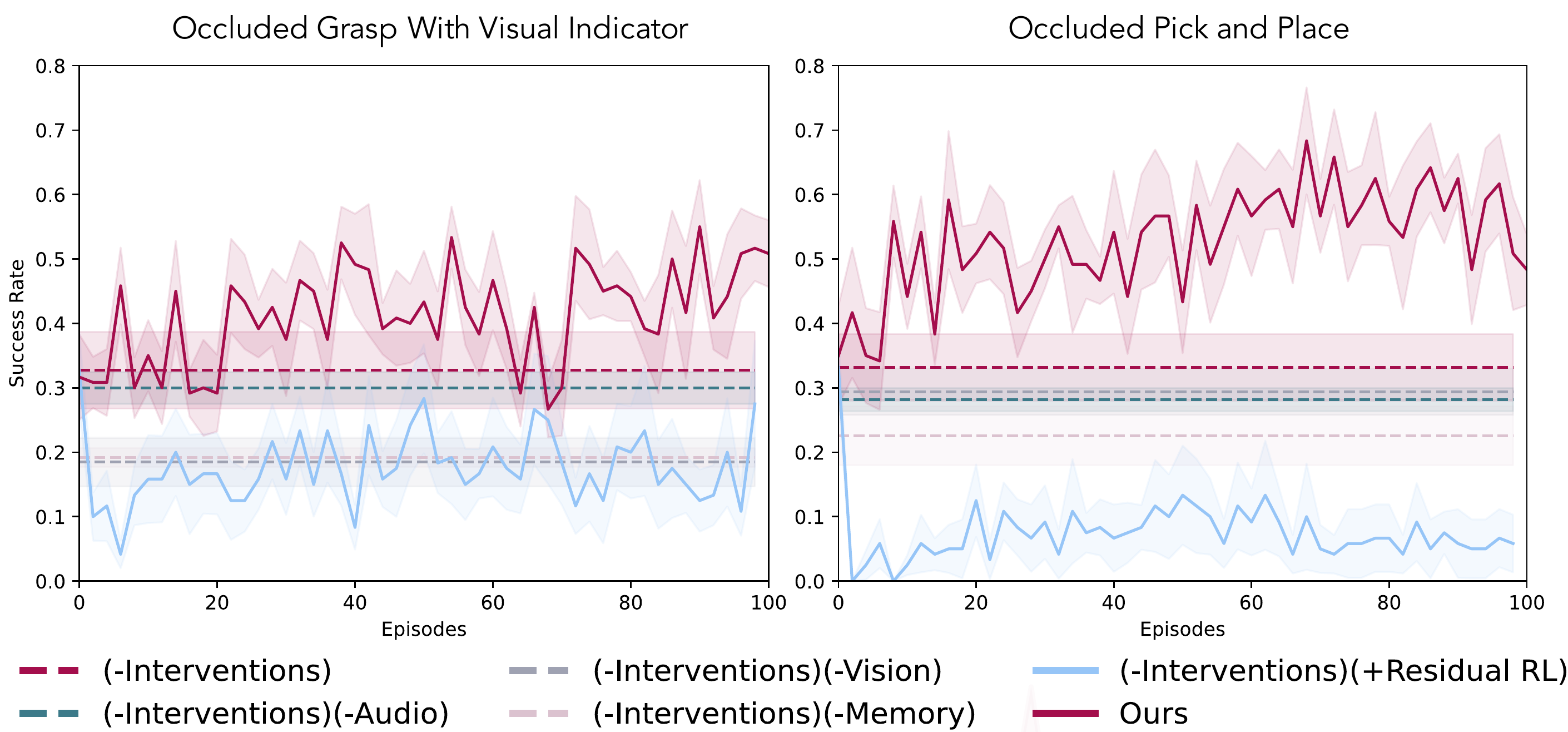}
    \vspace{-0.3cm}
    \caption{\small \textbf{Ablating System Components in Simulation}. We find that performance without online fine-tuning is considerably worse than our full method by $\sim\! 15\%$ in the grasp task and by $\sim\! 25\%$ on the pick and place task. In addition to removing interventions, removing vision, audio, or memory, further decreases the success rate of our method. Lastly we find residual reinforcement learning is unable to improve the performance of offline behavior cloning. Performance is reported as mean over 6 random seeds with standard error shading.}
    \vspace{-0.3cm}
    \label{exp1}
\end{figure*}

\begin{figure}[t]
    \centering
    \includegraphics[width=0.7\linewidth]{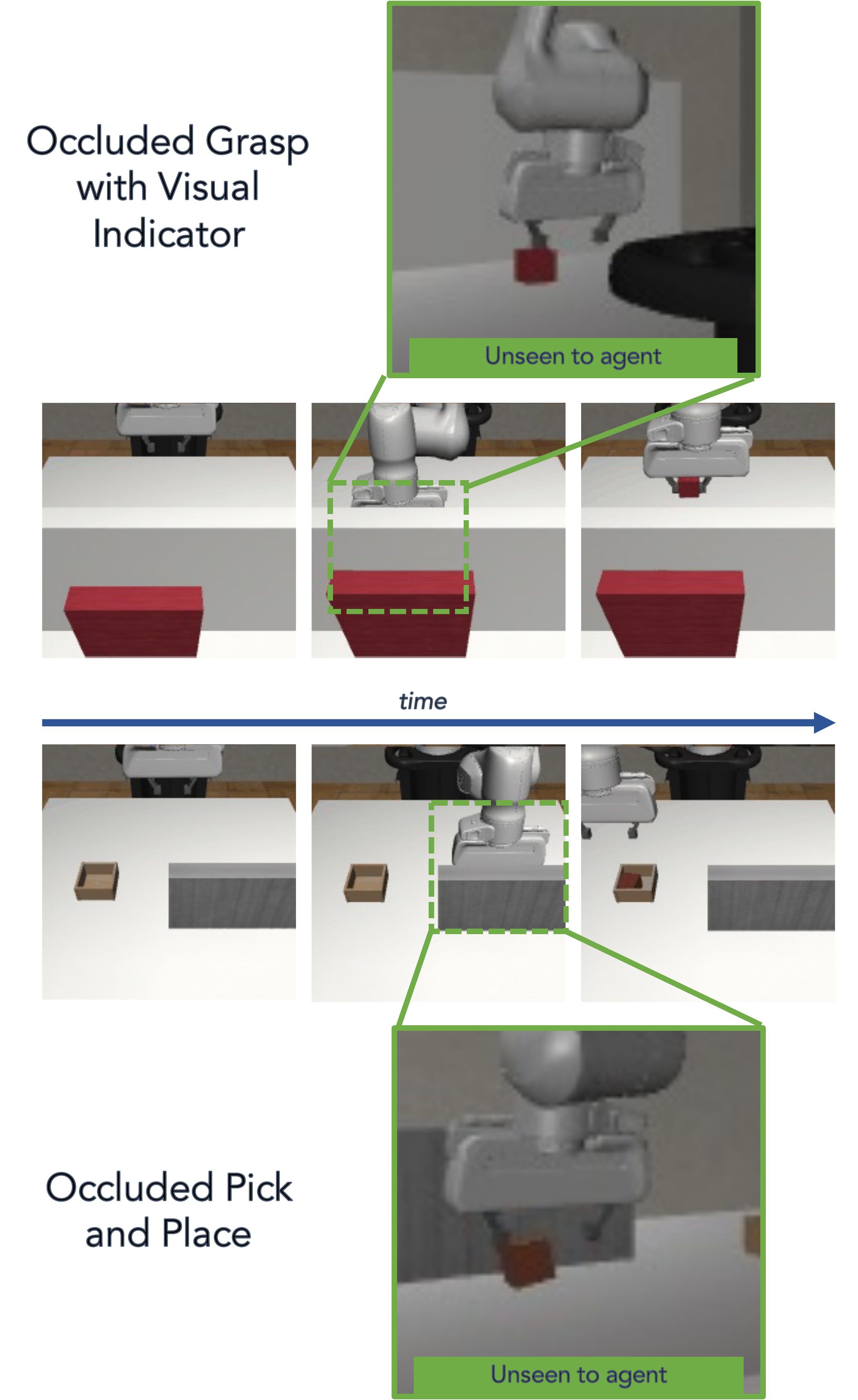}
    \vspace{-0.3cm}
    \caption{\small \textbf{Simulation Environments}. We design two simulation environments which require reasoning over vision and sound. Occluded Grasp with Visual Indicator (\textbf{top}) involves using vision and sound to localize a block behind a wall and grasp it. Occluded Pick and Place (\textbf{bottom}) requires using sound to localize and grasp a hidden block and vision to place it into a bin.}
    \vspace{-0.6cm}
    \label{simenv}
\end{figure}

\subsection{Can our Method Differentiate Between Object Sounds?}
\label{sec:robotsound}

In our previous experiment (Table~\ref{tab1}) we saw that when tested zero-shot on new objects, the policy extracted only those which made a sufficiently loud noise. This is expected, as the policy is only trained on data of one object, and thus only learned to detect a loud noise. But given training data of multiple objects, can our approach differentiate between them and use the sound to decide what actions to take next?

In our next experiment, we test exactly this on our real robot setup. Specifically, we design a task where the robot must first probe a bag of objects to listen for what object is inside the bag, and based on the object type, must either grasp the bag or stay in place. Specifically, the agent is trained to pick up and drop the bag, making a loud noise, and if the noise sounds like screws re-grasp it, otherwise if the noise sounds like packing peanuts, to leave it in place (See Figure~\ref{robotdomain2}). 

We train our model on 30 demonstrations, and evaluate it over 20 trials, 10 with the screws in the bag, and 10 with the packing peanuts. Our method succeeds in total \textbf{14/20} trials, 7/10 in each case. Thus, we conclude that giving training data of different objects, our method can differentiate between object sounds and choose actions accordingly. 

\subsection{Simulated Domain and Tasks}
\label{sec:sim}

Beyond our robot domain, we design two simulated environments to further test the systems ability to reason jointly over sound and vision in the presence of occlusion (See Figure~\ref{simenv}).

\vspace{0.1cm}\noindent\textbf{Occluded Grasp with Visual Indicator.} The robot's objective is to grasp a red cube that is randomly initialized behind a large wall which completely occludes the block. A visual indicator highlights which section behind the wall the block resides in, allowing the agent to find the block faster by using visual information. The optimal policy for this task would use vision to go behind the correct region of the wall, then search until a sound is made to localize the cube, and finally grasp the cube. A success is measured as the cube being lifted above a certain height within the horizon of the episode.

\vspace{0.1cm}\noindent\textbf{Occluded Pick and Place.} Like the previous task, the robot must grasp a red cube fully occluded behind a wall. However, in this task the cube is confined to a small region in the table, and the agent must not only grasp the cube, but also place it on the other side of the table in a bin (which is fully visible by the camera) whose position is randomized. Thus, the optimal policy for the task must use sound to localize and grasp the hidden block and use vision to place the block into the bin. A success is measured if the cube lands within a certain threshold of the center of the target bin. 

Both simulation environments are built in MuJoCo \cite{mujoco} on top of the \texttt{robosuite} \cite{robosuite2020} framework. In both environments the visual observations are 84x84 RGB images, and the action space is delta end-effector control and binary grasping. Since MuJoCo does not have sound simulation, we approximate sound with the external contact forces on the target object. 

We collect 50 demonstrations with a scripted expert that uses the ground truth simulation state. For online learning with interventions, we use a set of heuristics to determine when to intervene based on the ground truth state of the simulator and the timestep, and call the scripted expert policy for interventions. The offline policy is trained for 100,000 steps, and is finetuned online for 1,000 steps after each episode. 

\subsection{Ablating System Components in Simulation}
\label{sec:e2}

In simulation we further study the relative importance of the different components of our system. 
In Figure~\ref{exp1}, we observe that removing each component of the system hurts performance on both simulated tasks. 

First, performance without online fine-tuning \textbf{(-Interventions)} is considerably worse than our full method 
we observe that with interactive imitation by $\sim\! 15\%$ in the grasp task and by $\sim\! 25\%$ on the pick and place task.
Additionally, we find residual reinforcement learning \cite{Johannink2019ResidualRL} on top of the offline BC policy with SAC \cite{Haarnoja2018SoftAO} \textbf{(-Interventions)(+Residual RL)} to be ineffective, often destroying the performance of the policy trained with behavior cloning. We also run residual RL for 5x longer than our approach, and find that it still does not improve over the performance of offline BC. These results suggest that online learning from experts is critical to efficiently improving the success rate of the policy, matching our results in the real robot. 

Additionally, we ablate the use of each modality and memory from our method without online interventions. We see that on average removing memory \textbf{(-Interventions)(-Memory)} hurt performance the most, dropping performance by $\sim 10\%$. We find that removing audio \textbf{(-Interventions)(-Audio)} reduces the success rate by $\sim 5\%$ in both tasks, and removing vision \textbf{(-Interventions)(-Vision)} reduces the success rate by over 10\% in the occluded grasp task. 
Interestingly, we observe that in the simulator the agent's contact with the block can be perceived via a jolt in the movement of the robot arm, which is why the model without audio is able to achieve some success, albeit worse than our full approach. In the Occluded Grasp with Visual Indicator task, we see a bigger drop from removing vision, as the visual indicator is important in identifying the correct region to search for the block. On the Pick and Place task, since the target bin only moves around slightly, the agent is able to drop the block in the average position in the demonstrations and get a reasonable success rate, hence the limited drop in performance when removing vision.
Overall, we observe a consistent trend as on the robot that all components of our system are important to task success.

\section{Conclusion and Limitations}

We have proposed a system for tackling partially observed and heavily occluded manipulation tasks. Our key insights are to leverage sound, vision, and memory to resolve this partial observability, and to train our system in a data efficient way through interactive imitation learning. Our results on both the real robot and in simulation suggest that all components of our system are important to the final success rate. Ultimately, using our system we are able to accomplish up to 70\% success on the task of extracting keys from a bag from vision and sound. 

Despite this, a number of important limitations remain. First, our system depends heavily on human supervision in both offline demonstrations and in online human corrections. Rather than having a human always present, methods that can intelligently query the human \cite{Hoque2021ThriftyDAggerBN} could ease the burden on the human expert. Additionally, better techniques for finetuning through online reinforcement learning rather than expert feedback could make the learning process more autonomous. Second, while sound plays an important role in enabling the agent to learn when to grasp the keys, a single microphone only provides incomplete audio information. Much like humans have two ears, endowing the robot with two microphones could enable the robot to use sound in a more advanced capacity, e.g. to more precisely localize an object.

\section*{Acknowledgments}
The authors would like to thank Moo Jin Kim for assistance with the robot, and Samuel Clarke for valuable input on processing audio data. Toyota Research Institute provided funds to support this work.

\bibliographystyle{plainnat}
\bibliography{ref}

\newpage
\clearpage
\newpage 

\appendix
\section{Appendices}
\subsection{Architecture Details}
The models were created using the PyTorch \cite{pytorch} machine learning library. 

\textbf{Visual Encoder} We used an untrained Resnet-18 \cite{He2016DeepRL} encoder with the input modified to accept nine channels instead of three. The 512 dimensional encoding was extracted before the last fully-connected layer. 

\textbf{Spectrogram + Audio Encoder} Audio from the microphone is stored in a two-second rolling buffer at 48k samples per second. To preprocess, we apply a Fast Fourier Transform (FFT) with a segment length of 1920 samples. Because we wanted to demonstrate the robustness of the agent to extraneous audio, we did not apply any noise reduction algorithms. The FFT returns a $[57, 160]$ matrix, which we encode by using a 1D convolutional neural network. 

\textbf{LSTM} We combine the visual embedding with the audio embedding and proprioceptive data into one vector and fuse them through an MLP with two hidden layers of size $1024$ to yield a joint representation of size $50$. This joint representation is then fed into a one-layer LSTM for 10 time steps. 

\textbf{MLP} The LSTM outputs a temporal representation of size $50$, which is then fed into another MLP with two hidden layers of size $1024$ to yield the final action deltas. 

\textbf{Ablation Architectures} 
For visual ablations, we changed all image inputs to the policy into zeros. For audio ablations, we changed all audio signals to the policy into zeros. Finally, for memory ablations, we fed only 1 embedding vector into the LSTM (as opposed to the 10 in the full model), and remove frame stacking of the past 3 frames before encoding the visual observations. 

\subsection{Environment Details}

\textbf{Simulator} The two environments were created in MuJoCo \cite{mujoco} by modifying existing \texttt{Robosuite} \cite{robosuite2020} environments. We used a simulated Franka Panda robot, with position control through deltas at 20 Hz. A higher mass and friction were used by the block to prevent it from flying away on contact. Sound is approximated using the average external force on the object. 

\textbf{Simulated Expert Policy} The simulated expert policy uses the ground truth locations of the block (not shown to the real policy). A script monitors the ground truth location relative to the robot arm and outputs pre-determined actions depending on the state. For example, in the Occluded Pick and Place task, the script starts by moving until the cube is hit. It then lifts up enough to brush along the edge of the cube. After the gripper reaches a certain height, the script begins a grasping routine. Then, after the cube is firmly grasped, the script begins a lifting and moving routine. Finally, when the cube is over the bin (whose position is also known to the simulated expert policy), the script drops the cube. Stochasticity is added to the arm motion when appropriate. 

\textbf{Simulated Intervention Policy} Like the simulated expert policy, the intervention policy uses the ground truth location of the block relative to the robot arm and decides when to intervene. Interventions happen when the robot misses the block and keeps on moving past the block. It can also happen if the robot is close to the block but is not grasping it. Finally, interventions also happen when the robot drops the block after lifting it. When it intervenes, it uses the same simulated expert policy (described above) to correct the robot's actions. The intervention script also decides when to give control back to the robot to try again. This hand off of control happens after the script changes its behavior. For example, the expert policy might hand off control to the robot when it switches between grabbing and lifting. 

\textbf{Real Robot Environment} In our experiments we use a Franka-Emika panda robot placed behind a paper bag. The robot arm is controlled at 5 Hz with impedance-regulated delta commands. To reduce complexity, the orientation of the gripper is fixed. Inside the bag is a keychain with four metallic items of various size and shape. These items include (1) a standard lever-style bottle opener, (2) a screw cap bottle opener, (3) a souvenir medallion, and (4) a cylindrical whistle. Together, they provide consistent sound when moved around, but they also pose a significant challenge to the robot gripper. For example, if the gripper is not completely flush with the surface, it will not be able to grasp the cylindrical surface of the whistle. 

A third-person camera is placed in front of the robot / bag assembly. It is angled such that the robot and the top of the bag are visible but the keychain is not. This provides the occlusion needed for the experiment. 

A microphone is attached to the front of the gripper and makes direct contact with the surface of the gripper. This allows two mode of sound conduction. Sound can travel through the air to the microphone, but it can also travel through the robot gripper body. This combination allows for higher sensitivity, especially when the robot contacts the keys briefly. 

\textbf{Real Robot Expert Demonstrations} To collect the demonstrations on the real robot, we used an Oculus Quest VR system that tracked the position of the Oculus controller and relayed the deltas of the controller to the robot. In this way, when we move handheld controller, the robot moves in the same manner, with minimal delay. 

To facilitate the demo collection process, we mapped the front trigger of the Oculus controller to intiate a pause in demo recording. This allows us perform fine-grained motion by scaling down the deltas recorded by the controller without sacrificing the range. If any motions require more than an arm's length movement of the controller, a pause in demo recording allows us to reset the position of the controller before continuing. 

To control the gripper, we use the side trigger on the Oculus controller. We use a binary open/close command as reported by the state of the trigger. Due to problems encountered with rapidly changing gripper commands, we implemented a hard cooldown of ten timesteps (two seconds) per gripper command change. 

Finally, to terminate the episode early after a successful grasp and lift, we use the ``A'' button on the controller. Only successful demonstrations are saved in the replay buffer. 

\textbf{Real Robot Intervention Policy} Like the expert demos, the online interventions are done through the Oculus controller. Here, the front trigger is remapped to an override function. When the trigger is not pressed, the robot will act on its own policy. When the trigger is pressed, however, the deltas observed by the controller will override the robot's policy. These override events are recorded along with images, proprioception, and audio in the replay buffer, which allows for prioritized training. 

Early termination is also done after a successful intervention that leads to a grasp and lift. Unsuccessful episodes and episodes that require zero interventions are both discarded.

\subsection{Experiment/Training Details}

\textbf{Simulation Data Size and Training} In the simulation, we collected 50 demonstrations of 400 steps each. On this data we sampled batches of 16 transition sets and trained an imitation learning model to convergence, which was heuristically determined at 100000 steps. To reduce variation, 6 seeds were used to generate our plots. 

For the online interventions, we collect 100 intervention episodes and train in-between. In terms of wall time, the imitation learning takes around 10 hours on an RTX 2080 GPU, and the interventions take around 12 hours on the same hardware. 

\textbf{Simulation Evaluation} After training, we froze the weights and ran the robot for 250 evaluation episodes. For the Occluded Grasp environment, we defined a success as lifting the cube to a certain height above the table, such that the cube is visible from the camera. For the Occluded Pick and Place environment, we defined a success as the cube dropping into the bin such that its entire body is contained within the bin. 

\textbf{Real Robot Data Size and Training} Similarly, for the real robot we collected 50 demonstrations. As previously mentioned, we implemented early termination in the episode to account for greater variability in the grasp attempts. It takes on average between 35--70 steps for a successful grasp. This is comparable to simulation in terms of wall time, as the simulation control frequency is 4 times higher. 

However, because there was less data to train on, convergence during training happened at 10000 steps. We ran the imitation learning across three seeds and picked the best seed (through preliminary evaluations) to run online interventions on.

We collected 40 online interventions with training in-between and evaluations every 5 interventions. In terms of wall time, the imitation training takes around 1 hour on an RTX 2080 GPU, and the online intervention takes around 2.5 hours of supervised robot interaction. Most of this time was spent doing evaluations and the training in-between. The interventions take 15 seconds per episode. 

\textbf{Real Robot Evaluation}
At the end of imitation learning and for every 5 online intervention episodes, we run 20 evaluation episodes. In each episode, we give the robot 75 steps to successfully grasp and lift the keys into the field of view of the camera. We also allow for early termination if the robot succeeds before it finishes 75 steps or if it reaches a dangerous position (e.g. damaging the bag).  

\subsection{Additional Results}

\textbf{Full Failure Mode Analysis}

We find that, on the Easy version, our system's most common failure mode is \textit{unsuccessful lift} (3/3 failures) and, on the Hard version, it is \textit{no contact} with the keys (3/5) failures. For Ours (-Audio), the most common failure mode on Easy is \textit{false grasp} (8/8 failures) and on Hard is \textit{no contact} (7/10 failures). For Ours (-Vision), the most common failure mode on Easy is \textit{unsuccessful lift} (4/4 failures) and on Hard it is \textit{no contact} with the keys (5/8 failures). 
For Ours (-Memory), the most common failure mode on Easy is \textit{contact but no grasp} (6/6 failures) and on Hard it is \textit{no contact no search} with the keys (10/10 failures).

\end{document}